\documentclass{article}

\usepackage[english]{babel}

\usepackage[letterpaper,top=2cm,bottom=2cm,left=3cm,right=3cm,marginparwidth=1.75cm]{geometry}

\usepackage{amsmath}
\usepackage{graphicx}
\usepackage[colorlinks=true, allcolors=blue]{hyperref}
\usepackage{natbib}

\title{Augmenting Fundamental Analysis with Large Language Models: A RAG-Based System for Generating Investor Briefs}
\author{Bartosz Ziółko and Kacper Dobrzeniewski\\
Faculty of Computer Science, AGH University of Krakow, Poland\\
}

\begin{document}
\maketitle

\begin{abstract}
In this study, we examine the opportunities brought by Large Language Models (LLMs) to various aspects of fundamental analysis of companies based on their reports as well as data and documents describing macroeconomic situation like GDP and inflation changes as well as documents filled to the U.S. Securities and Exchange Commission (SEC) which can be found in EDGAR. We were preprocessing those data and than sending via API to gpt-4o model in a Retrieval-Augmented Generation (RAG) like regime. We prepared as well a document describing an exemplar investor knowledge based on Kitchin cycles. We were scanning data important for analysis of 9 companies for 4 weeks. Using LLM we were producing automatic briefs about them. They were sent to nine participants who are individual investors to evaluate usefulness of such approach to data analysis.
\end{abstract}



\section{Introduction}
Fundamental analysis, the cornerstone of investment strategies, traditionally relies on the in-depth evaluation of financial statements, industry reports, and macroeconomic indicators. This process, however, is time-consuming and requires analysts to process vast amounts of data, much of which is unstructured, such as the text from annual reports, earnings call transcripts, or documents filed with regulatory bodies like SEC via the EDGAR system. The rapid development in Natural Language Processing (NLP), and particularly the advent of LLMs, presents unprecedented opportunities to automate and deepen this process.

\subsection{Literature Review}

Early work on the application of NLP in finance primarily focused on sentiment analysis. The seminal study \cite{Loughran2011} demonstrated that general-purpose sentiment dictionaries are inadequate for the specific language of finance, leading to the development of specialized lexicons. With the advent of the Transformer architecture and models like BERT \cite{Devlin2019}, it became possible to create contextual models specifically pre-trained on financial corpora. An example is FinBERT \cite{Araci2019}, which significantly improved performance on sentiment classification tasks in financial texts.

The current revolution in NLP is fundamentally built upon the Transformer architecture, introduced by \cite{Vaswani2017}. Its core innovation, the self-attention mechanism, allows models to weigh the importance of different words in the input data, enabling a more profound contextual understanding and parallel processing capabilities that were not feasible with previous sequential architectures like Recurrent Neural Networks (RNNs) or Long Short-Term Memory (LSTM). This architectural breakthrough paved the way for the development of modern LLMs. Today, the field continues to advance at a rapid pace, with leading research labs pushing the boundaries of model capabilities. For instance, OpenAI's GPT-4o demonstrates a move towards natively multimodal models that can seamlessly process and reason across text, audio, and vision, increasing both efficiency and user interaction possibilities \cite{OpenAI2024gpt4o}. 

The emergence of large-scale generative models, such as those from the GPT family, has revolutionized the approach to text analysis. These models, with their ability to understand and generate text in a zero-shot or few-shot setting, can perform tasks far more complex than mere sentiment analysis. Research has shown their potential in forecasting financial markets. For instance, \cite{LopezLira2023} showed that ChatGPT can predict the direction of stock price movements from news headlines, suggesting these models internalize complex relationships between information and market reactions. In a similar vein, \cite{Itoh2023} explored the capabilities of ChatGPT for conducting textual analysis on fundamental data, further highlighting the potential of LLMs to automate and scale complex analytical tasks traditionally performed by humans.

One of the key challenges in applying LLMs in high-precision domains like finance is ensuring factual fidelity and avoiding so-called "hallucinations." The RAG approach, proposed by \cite{Lewis2020}, offers a solution to this challenge. It allows the model to "ground" its responses in specific, provided source documents, significantly increasing the reliability of the generated content, a benefit further substantiated by research showing that retrieval augmentation actively reduces model hallucination in conversational contexts \cite{Shuster2021}. Fundamentally, this process can be viewed as a practical implementation of controllable knowledge acquisition, where the LLM's "perception" of a situation is shaped by a curated set of inputs, a concept explored by \cite{Rzepka2023perception} as a cornerstone for building more reliable AI systems. Furthermore, the retrieval component of RAG can be enhanced beyond simple document fetching by employing structured knowledge bases, such as ontologies or knowledge graphs. This graph-based approach, as demonstrated by \cite{Rzepka2023} in the complex domain of security export controls, allows for more precise and context-aware information retrieval, paving the way for more sophisticated analytical systems. In parallel, to enhance analytical accuracy, models trained from scratch on financial data have also been developed, with a prime example being BloombergGPT \cite{Wu2023}, which demonstrated superior performance on a wide range of financial tasks compared to general-purpose models.

\subsection{Our Contribution}

Despite this rapid progress, the majority of existing research focuses on evaluating LLMs in isolated, specific tasks, such as price prediction, sentiment analysis, or the summarization of single documents. However, there is a lack of comprehensive studies that evaluate integrated systems that combine the analysis of company documents with macroeconomic data into a coherent and automated analytical workflow. Furthermore, the evaluation of such systems rarely extends beyond academic metrics to include an assessment of their \textbf{practical utility by end-users---individual investors}.

Our work fills this gap by presenting a system based on the $gpt-4o$ model within a RAG architecture. This system integrates data from diverse sources—company reports, macroeconomic data, and EDGAR filings—to generate synthetic analytical briefs. Additionally, the analytical context is enriched with expert knowledge regarding Kitchin business cycles. The key element of our study is the evaluation of the generated materials by a group of nine active individual investors, allowing for an assessment of the real-world value added by our approach in daily investment practice. This aligns with a growing body of research exploring how LLMs can be leveraged to systematically review and synthesize domain-specific knowledge, as demonstrated by \cite{Laniewski2024} in the context of algorithmic trading literature.

\section{Analysed data}

The selection of companies for this study was guided by a principle of \textbf{maximum diversity} to create a robust and challenging test environment for our LLM-based analytical system. The final cohort of nine companies was compiled based on several criteria, including market capitalization, sector, public visibility, and direct input from study participants. Each of the nine participants had the opportunity to nominate one company, resulting in a unique and varied portfolio.

The selected companies are: \textbf{Lululemon (LULU), NVIDIA (NVDA), Energy Fuels (UUUU), Coinbase (COIN), Corebridge Financial (CRBG), Pinterest (PINS), Novavax (NVAX), Amazon (AMZN), and Tesla (TSLA)}.

This selection methodology yielded a diverse set of firms, which serves the study's purpose by:
\begin{itemize}
    \item \textbf{Covering multiple sectors:} The portfolio includes technology (NVDA, PINS, AMZN, TSLA), consumer discretionary (LULU), financials (COIN, CRBG), uranium mining (UUUU), and biotechnology (NVAX).
    \item \textbf{Mixing company sizes and profiles:} It features large-cap, high-publicity stocks (like AMZN, NVDA, TSLA) alongside smaller or less-followed companies (like UUUU, NVAX, CRBG).
    \item \textbf{Testing adaptability:} This diversity ensures that our system was not overfitted to a single industry's jargon or reporting style and had to process a wide range of company-specific news, risks, and KPIs.
\end{itemize}

For this cohort of companies, we collected a comprehensive dataset over a four-week period. The data were structured to provide our RAG-based system with a multilayered context, spanning from high-level economic trends down to company-specific operational details.

\subsection{Macroeconomic and Cyclical Data}
This category provides a broad economic context. It includes key indicators for the U.S. economy as well as established cyclical and seasonal heuristics. These data were primarily sourced from official statistical releases and economic research platforms.
\begin{itemize}
    \item \textbf{Economic Indicators:} US GDP, US Inflation (CPI), US Services and Manufacturing PMI, Unemployment Rate, and Non-Farm Payrolls (NFP).
    \item \textbf{Monetary Policy:} Current and forecasted Federal Reserve funds rate.
    \item \textbf{Cyclical Data:} Kitchin Cycle, industry-specific cycles, the Presidential cycle, and tax cycles.
    \item \textbf{Seasonal Heuristics:} The "Sell in May and go away" anomaly.
\end{itemize}

\subsection{Company-Specific Fundamental Data}
This category forms the core of our analysis, focusing on data extracted directly from company reports (e.g. 10-K and 10-Q filings from EDGAR), press releases, and financial market data providers. We divided this into qualitative and quantitative subcategories.

\subsubsection{Qualitative and Operational Data}
These data are often textual and require interpretation, making them ideal candidates for analysis by our RAG-based system.
\begin{itemize}
    \item \textbf{Business Operations:} Significant geographic locations, key cost structures, and major customers.
    \item \textbf{Management \& Ownership:} Key shareholders, insider stock sales, and official management guidance or forecasts.
    \item \textbf{Company-Specific KPIs:} Industry-relevant Key Performance Indicators, such as:
    \begin{itemize}
        \item Comparable Sales, New Company-Operated Stores and Sale per $m^2$ for a store (LULU),
        \item Uranium spot price and Cameco and Kazatoprom production (UUUU),
        \item Bitcoin price, transaction volume and dates of court cases against SEC (COIN),
        \item Net Premiums Earned (NPE), Net Investment Income, Solvency Ratio, New Business Margin, Persistency Ratio and Claims Ratio (CRBG),
        \item Revenue Per User ARPU (PINS),
        \item Pipeline and Stage of Development, Cash-to-Operating Expenses Ratio and Price-to-sales (NVAX)
        \item AWS revenue and growth, Inventory Turnover Ratio, Customer Retention and Churn Rate and Advertising Revenue Growth (AMZN)
        \item Customer Retention and Churn Rate for FSD, production, delivery rate and revenue from energy sector (TSLA)
    \end{itemize}
\end{itemize}

\subsubsection{Financial and Market Data}
This includes quantitative metrics that reflect the company's financial health and market valuation.
\begin{itemize}
    \item \textbf{Valuation Metrics:} Price-to-Earnings (P/E) ratio and Forward P/E ratio.
    \item \textbf{Shareholder Returns:} Dividend payments and yield.
    \item \textbf{Market Sentiment:} Analyst recommendations.
    \item \textbf{Price Data:} Current stock price.
\end{itemize}

\section{Examples of Briefs Produced for Investors}

To provide the nine participants with timely and relevant insights, our system generated analytical briefs throughout the four-week study period. The generation process was dynamic and designed to mimic the information flow an active investor might face. Each week, the conceptual focus of the reports was slightly adjusted to test different aspects of the system's analytical capabilities.

Our daily operational strategy was twofold. Firstly, we prioritized generating briefs for the 2-3 companies experiencing the most significant activity on a given day, triggered by new data releases, significant news events, or sharp price movements. Secondly, we ensured a balanced coverage across the entire portfolio, guaranteeing that each of the nine selected companies received at least one dedicated brief per week, even in the absence of major news. This dual approach allowed us to test the system's reactivity to high-volume information flow as well as its ability to synthesize meaningful insights during quieter periods.

In this section, we present a selection of these briefs to illustrate the system's output. The examples have been chosen to demonstrate the variety of generated analyses, from event-driven reports to more routine weekly summaries. The biefs were originally produced in Polish apart from citition from news articles. We present translations.

\subsection{Tesla 15th November 2024 - Kitchin cycle analysis}

The generation of this brief for Tesla serves as a clear example of the system's core capabilities, particularly its capacity to synthesize information from diverse, unrelated sources into a coherent investment thesis. We analyse its performance across several key dimensions central to our research:

\begin{itemize}
    \item \textbf{Factual Grounding through RAG:} The system's RAG architecture proved effective in grounding the output in verifiable facts. The LLM successfully extracted specific financial figures (revenue of \$25.18 billion, P/E of 76.73) from their respective sources—a structured 10-Q report and a market data feed. Similarly, qualitative information, such as the names of insiders selling stock (Vaibhav Taneja, Kimbal Musk), was correctly identified, however it was preprocessed by us and it was not directly taken from unstructured SEC Form 4 filings. This demonstrates the model's ability to retrieve and accurately represent data from a heterogeneous knowledge base, which was a primary goal of the study.

    \item \textbf{Cross-Domain Synthesis:} A key success was the system's ability to connect macroeconomic context with company-specific analysis, a task central to fundamental investing. As instructed by the prompt, the system first executed a sub-task to determine the current economic phase. It correctly retrieved the latest PMI figures, interpreted their sub-50 values as indicative of a "slowdown" consistent with the Kitchin cycle framework, and then applied this context to evaluate the sectors Tesla operates in (Consumer Discretionary, Technology). This multi-step reasoning, combining cyclical analysis with corporate data, showcases a level of synthesis that goes beyond simple summarization.

    \item \textbf{Systematic Application of Heuristics:} The generation of qualitative judgments, such as labeling Tesla's financial position as "strong," was not left to the LLM's unconstrained discretion. The prompt contained a predefined heuristic template (e.g., "If revenue growth > X\% and debt/equity ratio < Y\%, describe the financial position as 'strong'"). This approach makes the analytical process more transparent, repeatable, and less susceptible to the LLM's potential biases, which is a crucial aspect of building a reliable analytical tool.

    \item \textbf{Human-in-the-Loop:} This particular brief required minimal post-processing. The generated text was coherent and factually accurate. Human intervention was limited to a minor grammatical correction in the summary and ensuring the formatting was consistent, validating the system's potential for a high degree of automation.
\end{itemize}

\itshape Tesla is a leader in electric vehicle production and autonomous driving technology, as well as a manufacturer of batteries and energy storage systems. The company is investing heavily in the development of charging infrastructure and renewable energy production, including solar roofs. Tesla's mission is to accelerate the transition to sustainable energy and transform the automotive industry toward electric power.

\#\#\# Reasons to buy TSLA stock:

1. **Strong financial position**: Tesla demonstrates solid financial performance, with growing revenues and profits. In Q3 2024, Tesla achieved revenue of \$25.18 billion, an 8\% increase year-over-year. Net profit was \$2.167 billion, also representing a 17\% increase year-over-year. A strong cash position (\$33.648 billion) and low debt (debt/equity ratio of 18\%) are additional advantages.

2. **Energy Segment Growth**: Tesla continues to expand its energy sector, achieving record gross margins in this segment. Energy revenue increased 52\% year-over-year, demonstrating growth potential in this area.

3. **Innovation and Technological Development**: Tesla is investing in AI and autonomous driving technologies, which could yield long-term benefits. Product developments like the Cybertruck and new vehicle models based on the next-generation platform also point to future growth.

4. **Strategic Policy Approach**: Tesla could benefit from the planned elimination of tax breaks for electric vehicles in the US, which could harm competitors more than Tesla itself. Elon Musk believes that the elimination of tax breaks could benefit Tesla in the long term.

\#\#\# Arguments for selling TSLA stock:

1. **High Valuation**: Tesla's current P/E is 76.73, which may suggest the stock is overvalued, increasing the risk of a price correction.

2. **Regulatory Uncertainty**: The Trump Administration's planned elimination of tax breaks for electric vehicles could negatively impact Tesla's sales in the short term, although it could harm competitors more in the long term.

3. **Macroeconomic Changes**: Current data suggests the US economy is in a slowdown, which could negatively impact the consumer discretionary and technology sectors, which Tesla belongs to.

4. **Insider Stock Sales**: Stock sales by key insiders, such as CFO Vaibhav Taneja and Kimbal Musk, could signal that company insiders anticipate potential difficulties or a price correction.

\#\#\# Economic Phase:
Based on available data, we are currently in an economic slowdown phase. This is indicated by the Industrial PMI and ISM Manufacturing PMI falling below 50, suggesting an economic slowdown. During this phase, the energy, real estate, consumer staples, and utilities sectors are the best investments, while consumer discretionary and technology sectors may perform less well.

\#\#\# Summary:
The decision to buy or sell Tesla stock should consider both the company's strong financial fundamentals and innovative strength, as well as the potential risks associated with its high valuation and regulatory uncertainty. Investors should also consider the current macroeconomic conditions and the phase of the economic cycle we are in.\normalfont

\subsection{NVIDIA 19th November 2024 - news analysis}
This example demonstrates a different core capability of our system: the processing, summarization,
and ranking of news articles based on the perspective of an investor observing the market.
The input consisted of numerous articles concerning NVIDIA published on November 18 and 19. The first limitation of this study was the token limit imposed by OpenAI. The length of individual articles frequently exceeded the model's capacity within a single prompt, effectively hindering the possibility of conducting a comprehensive analysis. Therefore, the experiment needed to be divided into phases. The initial step involved summarizing all the articles by the model. Subsequently, as the second step, these summaries were fed into a RAG prompt along with an instruction to rank them according to their level of importance (without additional context provided). The results strongly favored political content over economic content. Nonetheless, the objective of the study was to determine which articles are reliable as they have an impact on stock prices and could provide a basis for recommendations or serve as strong indications for the language model in RAG. 
\subsubsection{Ranking}
A necessary step in the methodology was to construct a ranking framework for the model. This framework prioritizes news sources according to the following descending order of importance:
\begin{enumerate}
\item Corporate quarterly and annual reports.
\item Decisions and announcements from central banks (e.g., the Federal Reserve).
\item Macroeconomic data, including indicators on inflation, employment, and economic activity.
\item Major corporate events such as mergers and acquisitions (M\&A), restructurings, strategic announcements, and changes in executive management.
\item Insider trading activity and transactions by large institutional investors.
\item Reports from competitors or other companies within the same industry sector.
\item News articles, with a focus on their sentiment, novelty, and topicality.
\item Analyst reports, including ratings and price targets.
\end{enumerate}
Another challenge was the number of news articles that focused on specific companies. The volume of articles narrowed the time horizon to a single day for the most popular companies, which significantly affected the organization of the news, effectively reversing the order of the aforementioned list, prioritize sentiment, novelty, ratings, and analysts. To balance this, the latest financial report of the respective company was recently added into the RAG, with the recommendation that data from this report be taken into account when compiling the list of news articles.
\subsubsection{Ranked News Summaries (System Output)}
Below is an example of summarized articles ranked by their significance to investors for the company NVIDIA on November 18 and November 19, up until the market opening hours:
\itshape \begin{enumerate}
    \item (Valuation Milestone): Prediction that NVIDIA may become the first company in the world to reach a valuation of \$4 trillion, approaching this milestone with a current market capitalization of approximately \$3.6 trillion. NVIDIA's growth is driven by increasing demand for its graphics processors, particularly in the context of artificial intelligence, which may lead to further revenue and profit expansion.
    \item (Technical Risk): Reports on overheating issues affecting NVIDIA’s Blackwell chips, which have contributed to a decline in the company’s stock price ahead of the upcoming quarterly earnings report. Analysts suggest that, despite the potential loss of market share, NVIDIA may still strengthen its position in the AI sector.
    \item (Earnings Preview): Discussion of high expectations surrounding NVIDIA’s quarterly results, pointing to a potential stock price movement of nearly \$300 billion. It also outlines reasons why investors might consider either buying or holding off on purchasing shares before the earnings announcement. While strong demand for NVIDIA’s products is noted, concerns about the new Blackwell chips could impact financial outcomes and market response.
    \item (Technical Risk - Confirmation): NVIDIA’s stock declined due to concerns about overheating in its Blackwell processors, potentially affecting the company’s ongoing development. Issues with the new AI chips may negatively influence investor confidence and NVIDIA’s future financial performance.
    \item (Bullish Counterpoint): The article predicts that NVIDIA’s shares will rise following the earnings announcement on November 20, primarily due to strong demand for the new Blackwell processors, upgraded profit forecasts by analysts, and expected increases in the company’s clients’ capital expenditures. These factors could significantly contribute to the growth of NVIDIA’s stock value in the coming weeks.
    \item (Geopolitical/Ecosystem): Reports on Nebius Group plans to launch its first cloud operations in the United States, which may boost the use of NVIDIA chips in AI, against the backdrop of rising geopolitical tensions related to the war in Ukraine affecting stock markets. In light of these tensions, investors are concerned about market stability, leading to declines in indices such as the Dow, S\&P 500, and Nasdaq.
    \item (Partnership News): Foxconn has partnered with NVIDIA to implement digital twin technology in its factories, aiming to streamline production processes and supply chain management while increasing the company’s resilience to disruptions. This partnership has the potential to revolutionize global manufacturing operations and enhance Foxconn’s competitiveness in global markets.
    \item (Analyst Commentary): Jim Cramer warns NVIDIA investors against overreacting to the latest Blackwell report, suggesting that the stock decline is natural and should not cause panic. He emphasizes that the company’s long-term fundamentals remain strong, which should reassure investors.
    \item (Investor View - Bullish): The article discusses Masayoshi Son’s views on NVIDIA Corporation, which he considers undervalued. According to Son, spending on artificial intelligence is expected to increase significantly over the coming years, benefiting NVIDIA’s GPU-related business. Furthermore, reports indicate growing investor interest in AI startups, which raised nearly \$19 billion in funding in the third quarter of 2024.
    \item (Investor View - Bearish): Philippe Laffont, billionaire and founder of Coatue Management, sold 80\% of his stake in NVIDIA, reallocating investments to Alibaba, which he considers undervalued in the AI context. Laffont sees Alibaba’s strong position in China’s e-commerce market and its expanding cloud segment as factors making the company an attractive investment.
    \item (Investor View - Bearish): The article covers the investment decisions of billionaire Stanley Druckenmiller, who sold all his shares in NVIDIA, considering their valuation too high, and invested in Broadcom, a company with growing potential in artificial intelligence. This indicates his belief in the future development of the AI market alongside a cautious approach to investing in highly valued stocks.
    \item (Competitor News): AMD announced a 4\% workforce reduction affecting about 1,000 employees, despite achieving over one billion dollars in profit in the third quarter of 2023 and growth in the AI sector. This decision raises questions as to whether it is merely a resource cut or a sign of deeper financial issues within the company.
    \item (Supply Chain/Ecosystem): Samsung Electronics plans to secure orders from Qualcomm and NVIDIA for 2nm and 3nm processes, while simultaneously developing its own Exynos processors despite production performance challenges. The company is striving to increase self-sufficiency in application processor manufacturing, which could influence the future of its smartphone segment strategy.    
\end{enumerate}
\normalfont

\subsubsection{System Performance and Analysis}

This news-ranking task provides valuable insights into the LLM's ability to reason under conditions of information overload and conflicting reports.
\begin{itemize}
    \item \textbf{Prioritization of Market-Moving Themes:} The system correctly identified the three dominant narratives for NVIDIA on this day: (1) the massive, forward-looking valuation potential, (2) the immediate, tangible risk of technical issues with a key new product (Blackwell), and (3) the imminent, high-impact binary event (the earnings report). These themes occupy the top 5 spots, demonstrating a sophisticated understanding of what truly matters to investors.
    
    \item \textbf{Identification of Conflicting Information:} The model did not simply rank all positive news first. It placed a negative report about overheating chips at \#2, immediately after a positive report on valuation. This highlights the system's ability to recognize the high significance of a potential risk, even amidst bullish long-term narratives. It also correctly clustered a bullish counterpoint on the same topic at rank \#5.

    \item \textbf{Contextual Discounting of Information:} The ranking demonstrates a logical hierarchy. Direct news about NVIDIA's valuation, products, and financials was ranked highest. Commentaries from individual analysts or investors (Cramer, Son, Laffont, Druckenmiller) were ranked in the middle tier. Broader ecosystem or competitor news (Foxconn, AMD, Samsung) was correctly placed at the bottom, recognized as relevant but less immediately impactful than core company issues. This shows the model successfully applied the nuanced ranking framework described in our methodology.

    \item \textbf{Challenge of Nuance:} While the ranking is logical, the model's summaries do not fully capture the nuance of the bearish investor sentiment (Laffont, Druckenmiller). Both investors sold not because they believe NVIDIA is a bad company, but because of its high valuation—a crucial distinction. This highlights a limitation and an area for future improvement in prompting the model to extract not just the action (selling) but the detailed rationale behind it.
\end{itemize}


\subsection{Revenue Exposure and Geopolitical Risk: NVIDIA, Amazon, and Tesla - 22nd November 2024}

LLM models can also be used in the context of extracting data from attached text in RAG. An example of this could be research based on quarterly reports, focusing on the information they contain about risks, such as how the unavailability of the Chinese and Taiwanese markets could affect a given company. The following are examples for companies: NVIDIA, Amazon, and Tesla. All of this information is based on the quarterly Q3 2024 reports.

\subsubsection{NVIDIA}
\itshape
Based on the provided financial data, NVIDIA’s revenue from China (including Hong Kong) and Taiwan for the nine months ending October 27, 2024, amounted to \$11.574 billion and \$15.266 billion, respectively. This represents a combined total of \$26.84 billion out of \$91.166 billion in total revenue for the same period. Consequently, should access to the Chinese and Taiwanese markets be disrupted due to war or sanctions, NVIDIA could potentially lose approximately 29.44\% of its revenue, assuming no alternative markets or compensatory sales strategies are implemented. Such a disruption would likely have a significant impact on both the company’s production and sales, given the substantial contribution of these regions to its total revenue.

\textbf{Revenue from China:}
The report notes that NVIDIA’s Data Center segment revenue from China increased in the third quarter of fiscal year 2025 compared to the previous quarter. However, as a percentage of total Data Center revenue, it remained below the levels recorded prior to the implementation of export controls in October 2023. This suggests that China remains a significant market for NVIDIA’s Data Center products, although the exact share of revenue derived from China is not specified.

\textbf{Export Controls and Licensing Requirements:}
The text also addresses various export controls and licensing requirements imposed by the U.S. government, which restrict NVIDIA’s ability to sell certain advanced products in China. These limitations indicate that a portion of NVIDIA’s revenue is contingent on its capacity to export to the Chinese market. The imposed restrictions have already affected the company’s competitive positioning.

\textbf{Geopolitical Risks:}
The report emphasizes geopolitical tensions—particularly those involving Taiwan and China—as potential threats that could severely disrupt NVIDIA’s supply chain and revenue streams. Taiwan plays a critical role in NVIDIA’s supply chain, and any disruptions in this region may significantly affect the company’s manufacturing capabilities. The overall impact would likely be substantial, particularly in the Data Center segment and across the supply chain. However, the precise extent would depend on numerous factors, including the company’s ability to redirect sales and restructure its supply chain.
\normalfont

\subsubsection{Amazon}
\itshape
Amazon’s International segment accounted for 23\% of its consolidated revenue in the third quarter of 2024. Should access to the Chinese and Taiwanese markets be lost due to war or sanctions, there could be a notable impact on Amazon’s international revenue. However, specific figures for revenue generated solely from China and Taiwan are not provided, making it impossible to determine the exact financial impact based solely on the available information. The total effect would depend on the proportion of international sales attributable to these two markets.
\normalfont
\subsubsection{Tesla}
\itshape
Tesla reported the following revenue from the Chinese market:
\begin{itemize}
    \item Three months ending September 30, 2024: \$5.665 billion
    \item Nine months ending September 30, 2024: \$14.893 billion
\end{itemize}

If access to the Chinese and Taiwanese markets were lost, Tesla would forgo this revenue. Since no specific figures for Taiwan are provided, it is assumed that any impact from that region is included within the "Other international" category. Based on this assumption, the loss of the Chinese market would result in a revenue decline of approximately 20.7\% for the nine-month period ending September 30, 2024 (\$14.893 billion out of \$71.983 billion in total revenue).

Moreover, production at the Gigafactory Shanghai - which manufactures Models 3 and Y - would also have to be suspended, which would affect Tesla’s global production capacity. However, the text does not provide detailed estimates regarding the production impact, and therefore a precise assessment cannot be made without further data.
\normalfont

\subsection{Energy Fuels 29th November 2024 - Fundamental Analysis of a Small-Cap Uranium Company}

The analysis of Energy Fuels showcases the system's proficiency in several advanced tasks that go beyond simple news summarization and are critical for evaluating smaller companies where data is often less structured and more concentrated.

\begin{itemize}
    \item \textbf{Deep Extraction from a Single Source:} Unlike the NVIDIA example which processed many articles, this brief was primarily derived from one key press release. The LLM demonstrated its ability to dissect a complex corporate document, correctly identifying and separating distinct positive developments (new uranium contracts, REE production milestones, the lifting of the Toliara project suspension) and structuring them into a coherent "News Summary" and distinct "Buy Reasons."

    \item \textbf{Identification of Contradictory Signals:} A significant success of the system was its ability to maintain analytical balance. Despite the overwhelmingly positive operational news in the press release, the RAG process also retrieved the headline financial figure—a net loss of \$12 million—from the attached financial statements. The system correctly flagged this as a primary "Argument for Sale," demonstrating its capacity to identify and present conflicting information, which is a hallmark of robust financial analysis.

    \item \textbf{Prompt-Guided Financial Calculation:} The "Financial Analysis" subsection highlights a more advanced capability. The prompt contained instructions for the LLM to perform basic, yet insightful, calculations based on extracted data. The system correctly retrieved the cash position (\$47.46M), market capitalization (\$1.2B), and quarterly loss (\$12M), and then successfully calculated the cash-to-capitalization ratio and a basic "cash burn" runway ("survive for about four quarters"). This demonstrates a move from qualitative summarization to quantitative, heuristic-based analysis.

    \item \textbf{Nuanced Macroeconomic Application:} The system once again applied the macroeconomic framework, but with a specific nuance relevant to this company. It correctly identified that while a recessionary phase is generally negative, it can be beneficial for specific sectors like energy. Furthermore, it made a sophisticated connection between the macroeconomic trend of falling interest rates and the company's specific financial state (no debt), correctly identifying this as a potential advantage.
    
    \item \textbf{Creative Synthesis in Titling:} The brief's title—"UUUU: Uranium Turmoil and Rare Earth Opportunities - Is This Your Goldmine?"—was generated by a specific instruction in the prompt to create an engaging, journalistic-style headline. This showcases the model's flexibility to adapt its output style, moving from purely technical prose to more creative and accessible language suitable for an investor audience.
\end{itemize}

\itshape\#\#\# Title: "UUUU: Uranium Turmoil and Rare Earth Opportunities - Is This Your Goldmine?"

Energy Fuels is an American company engaged in the mining and processing of uranium and vanadium, key raw materials for the nuclear energy industry. The company also has resources and technologies for the extraction of rare earth elements, which are used in advanced technology industries. Its operations are focused on the sustainable development of energy resources, primarily in the United States.

\#\#\# News Summary:
Energy Fuels announced its Q3 2024 results, highlighting active uranium mining and processing operations and successes in rare earth element (REE) production. The company continues to build a REE supply chain hub in the US. Key achievements include a new uranium sales contract and the commencement of processing of a large uranium inventory at the White Mesa Mill. Additionally, the company has completed the acquisition of Base Resources, which is expected to accelerate the development of its heavy mineral sands and REE projects, including the Toliara project in Madagascar, following the lifting of the suspension by the Madagascar government.
https://www.prnewswire.com/news-releases/madagascar-government-lifts-suspension-on-energy-fuels-toliara-critical-minerals-project-302318407.html

\#\#\# Buy Reasons:

1. **Strong Financial Position**: Energy Fuels has a solid financial position with over \$180 million in liquidity and no debt, providing stability in the face of changing market conditions.

2. **Uranium as a Key Product**: The company has secured a new long-term contract for uranium supplies to the US, ensuring stable revenues in the coming years. 

3. **REE Development**: Energy Fuels has successfully launched an REE separation circuit, which could attract customers from the technology and defense industries.

4. **Lift of the Toliara Project Suspension**: The Madagascar government's decision to lift the suspension allows for the development of a project that could become a significant revenue source.

5. **Projected REE Market Growth**: Growing demand for rare earth elements in the technology and automotive sectors supports the company's long-term growth.

\#\#\# Arguments for Sale:

1. **Financial Loss**: The company reported a net loss of \$12 million, which may raise investor concerns about profitability.

2. **High Operating Costs**: Mining and processing uranium and REE are expensive, and environmental regulations may increase costs. 

3. **Downward Uranium Price Trend**: Current uranium prices are trending downward, which could impact profit margins.

4. **Macroeconomic Uncertainty**: Current economic conditions suggest a slowdown, which could impact commodity demand.

\#\#\# Financial Analysis:

- **Cash vs. Capitalization**: Energy Fuels has \$47.46 million in cash with a market capitalization of \$1.2 billion, resulting in a cash-to-capitalization ratio of approximately 4\%, which is rather low.

- **Survival on a Loss**: With a quarterly loss of \$12 million, the company could survive for about four quarters without additional financing.

\#\#\# Macroeconomic Impact:
The current economic phase suggests a recession, which could boost investments in energy-related sectors such as uranium. Falling interest rates could also favor investments in debt-free companies like Energy Fuels.

\#\#\# Summary:
Investing in Energy Fuels may be attractive to those who believe in the long-term growth of the REE market and the company's financial stability, but potential risks related to operating costs and trends in uranium prices should be closely monitored.\normalfont

\subsection{Evaluating LLM Responses to Unforeseen Events: The Case of Martial Law in South Korea - 4th December 2024}

The study also examined how LLM responds to highly unexpected events that are difficult to predict and quantify in a quarterly report, specifically, the imposition of martial law in South Korea. To recall, martial law in South Korea was declared on December 3, 2024, at 10:27 p.m. local time, by President Yoon Suk-yeol during a live televised address. The state of martial law lasted approximately six hours and was lifted on 4 December 2024, at around 4:30 a.m., following a vote in the National Assembly. The following are examples for Coinbase, NVIDIA, and Lululemon.

\subsubsection{Coinbase}
\itshape
The imposition of martial law in South Korea may significantly affect the operations of a publicly traded company, especially one engaged in international business and operating in highly regulated sectors such as cryptocurrencies, mobile technologies, or financial services.

\textbf{Heightened Regulatory Risk:}
Martial law may lead to stricter regulation and government oversight of business operations, particularly in areas related to anti-money laundering (AML), counter-terrorism financing, and compliance with international sanctions. The company may be required to rapidly adjust to new legal requirements, which could incur additional costs and expose it to legal risks.

\textbf{Reputational Impact:}
If the company is perceived to have connections to North Korea or other sanctioned entities, its reputation could suffer. Increased media and regulatory scrutiny can erode investor and customer confidence.

\textbf{Financial Risk:}
Potential sanctions or operational restrictions may impair the company’s ability to conduct business in specific regions, resulting in revenue loss. In addition, financial penalties for compliance violations could further burden the company’s balance sheet.

\textbf{Stock Price Volatility}
Depending on the perception of the investor of the regulatory and political risks, the company’s stock price could experience significant volatility. If risks are deemed high, the stock may decline. Conversely, if the company demonstrates effective risk management and operational adaptability, market response may be more favorable.

In summary, the imposition of martial law in South Korea could increase both regulatory and operational uncertainty for publicly traded companies, potentially reducing their market valuation unless effective risk mitigation strategies are deployed.
\normalfont
\subsubsection{NVIDIA}
\itshape
The introduction of martial law in South Korea could have a significant impact on the company’s operations. Political instability of this magnitude may elevate operational risks and investor concerns, potentially affecting the firm’s market valuation.

\textbf{Increased Regulatory and Political Risk}
Martial law may lead to the tightening of laws and regulations, increasing compliance costs, and the likelihood of violations that could result in both financial penalties and reputational damage.

\textbf{Supply Chain Disruptions}
If the company maintains suppliers or clients in South Korea, martial law could disrupt its supply chain. Such disruptions may cause delays in production and delivery, adversely impacting financial performance.

\textbf{Investor Sentiment}
Investors may view the company as a riskier investment due to the unstable political environment, potentially triggering a decline in share price.

\textbf{Potential Sanctions and Trade Restrictions}
Should the political situation escalate further, international sanctions or trade restrictions could be imposed, affecting the company’s ability to operate in global markets.

In conclusion, the imposition of martial law in South Korea is likely to negatively affect the company’s stock price due to heightened regulatory risks, supply chain challenges, and adverse investor sentiment.
\normalfont
\subsubsection{Lululemon}
\itshape
Martial law in South Korea could have a material effect on the operations and market valuation of publicly traded companies. In the case of the company analyzed in the report, which operates 19 retail stores in South Korea, such a political event could lead to store closures, supply chain interruptions, a decrease in foot traffic and general economic uncertainty in the region. These factors could result in a reduction in revenue and profits derived from the Korean market, potentially affecting the company's stock price.

However, it is worth noting that the company operates 19 stores in South Korea out of a global total of 721, meaning Korean locations account for approximately 2.64\% of its overall store count. Assuming revenue is roughly proportional to the number of stores, it may be inferred that the Korean market contributes a similar percentage to the company’s total revenue. Nevertheless, the actual impact may vary depending on local market conditions, brand popularity, and other region-specific factors.
\normalfont

\subsection{Lululemon 6th December 2024 - clusterization of news articles}

This example demonstrates a more advanced NLP task performed by our system: the automated clustering and thematic analysis of news articles following a major corporate event. In situations of high information volume, such as after a quarterly earnings release, merely summarizing or ranking articles can be insufficient. The goal here was to test the system's ability to first identify the distinct underlying narratives present in the news flow, and then to represent each narrative both textually and visually.

The system used LanceDB \cite{She2024LanceDB} and K-means \cite{MacQueen1967} algorithm to process all news articles about Lululemon published after its Q3 2024 earnings announcement. It grouped semantically similar articles into clusters, generated a summary for each cluster. We also created a visual "word map" to highlight the core themes of each narrative using WordCloud library \cite{MuellerWordCloud}.

This clustering task reveals the system's capability to move beyond linear analysis (ranking) to a more sophisticated, multi-faceted understanding of information landscapes.

\begin{figure}
\includegraphics[width=\textwidth]{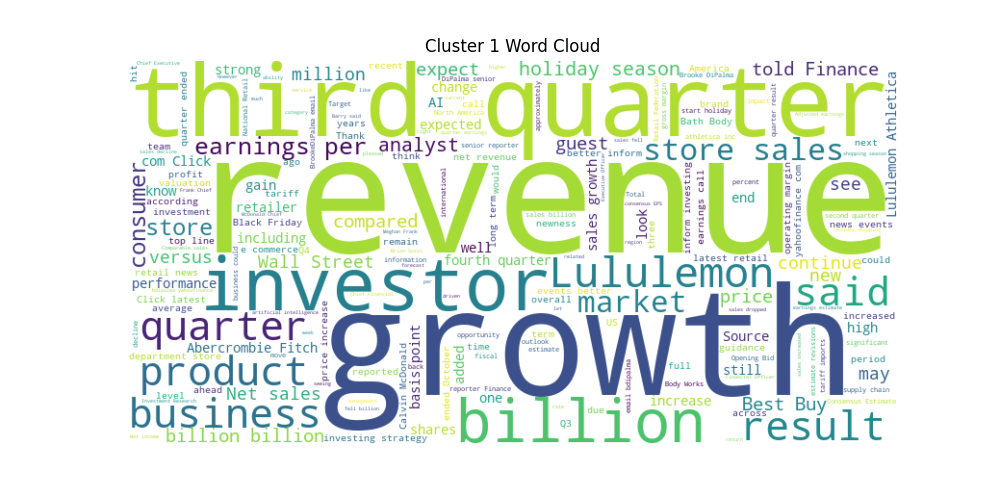}
\caption{Word map used in the brief representing a cluster of similar news articles about Lululemon on 5th and begining of 6th of December}
\label{LULU1}
\end{figure}

\begin{figure}
\includegraphics[width=\textwidth]{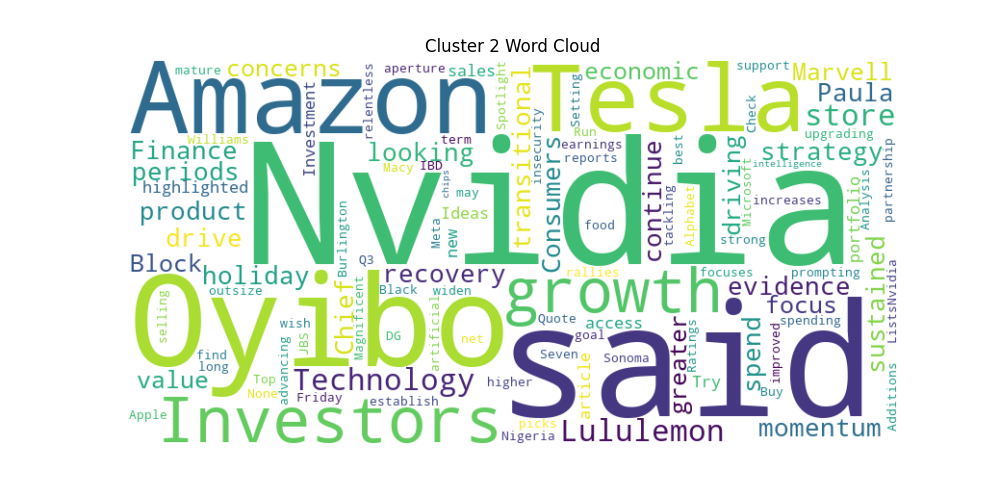}
\caption{Word map used in the brief representing a cluster of similar news articles about Lululemon on 5th and begining of 6th of December}
\label{LULU2}
\end{figure}

\begin{itemize}
    \item \textbf{Automated Narrative Identification:} The system successfully performed an unsupervised clustering task, identifying two primary, distinct narratives from the pool of post-earnings articles. This demonstrates an ability to discern not just what is being said, but the different angles from which the event is being reported. This is a significant step beyond simple sentiment analysis.

    \item \textbf{Cluster 1: The Factual, Data-Driven Narrative:} The first cluster (illustrated by Fig. \ref{LULU1}) correctly grouped all articles focused on the direct, quantitative results of the earnings report. The LLM's summary accurately extracted the key data points: the beat on revenue and EPS, the raised full-year forecast, and the specific counterpoint of slowing sales growth in North America. The word map for this cluster would visually reinforce the dominance of terms like "revenue," "earnings," "beat," "forecast," and "North America."

    \item \textbf{Cluster 2: The Forward-Looking, Strategic Narrative:} The second cluster (illustrated by Fig. \ref{LULU2}) showcases a more nuanced capability. The system isolated a different type of article—one that placed the immediate results in a broader, forward-looking strategic context. It successfully extracted qualitative insights, such as the CFO's description of the upcoming year as a "transitional period" and the observation of a consumer shift towards "value." This demonstrates the system's ability to separate tactical, data-driven reporting from strategic, forward-looking commentary.
    
    \item \textbf{Handling Irrelevant Data:} It is noteworthy that the source articles for the second cluster contained tangential information about other companies (e.g., Dollar General). The LLM's summary correctly filtered out this irrelevant noise and focused only on the content pertinent to Lululemon, showcasing its robustness in real-world, imperfect data environments.

    \item \textbf{Data Visualization as a Synthesis Tool:} The generation of word maps is a key feature of this module. This visual output serves as an instant, intuitive summary of a cluster's core themes. It allows an investor to grasp the essence of a narrative "at a glance," providing a powerful cognitive shortcut that complements the detailed textual summary. This highlights the system's potential not just as an analytical engine, but also as a tool for effective information visualization.
\end{itemize}

\itshape Lululemon Athletica is a global premium brand specializing in the production and sale of sportswear and accessories, particularly valued in the athleisure segment. The company stands out for its strong commitment to quality, innovative materials, and loyal customer base, which allows it to achieve high margins and dynamic sales growth. Strategic investments in e-commerce and international expansion, as well as the development of its men's line and digital products, strengthen Lululemon's position as a leader in the sportswear and lifestyle apparel sector.

Lululemon (LULU) reported third quarter results after the closing bell on Thursday that beat on both the top and bottom lines, sending shares of the company higher in after-hours trade. Lululemon stock rose over 8\% as the company also raised its full-year sales and profit forecasts for 2024. Still, sales growth in North America once again declined as the retailer grapples with concerns over increased competition heading into the critical holiday shopping season. Revenue came in at \$2.40 billion, an increase from the \$2.20 billion reported in the third quarter of 2023. Analysts polled by Bloomberg were expecting \$2.36 billion after the retailer guided to sales between \$2.34 billion and \$2.37 billion. Earnings beat estimates of \$2.75 a share to hit \$2.87. This was also ahead of the \$2.53 EPS the company reported in the year-ago period. The company guided to fourth quarter revenue of \$3.48 billion-\$3.51 billion, compared to consensus estimates of \$3.5 billion. […]
https://finance.yahoo.com/news/lululemon-stock-rises-on-profit-beat-as-company-boosts-full-year-outlook-211820899.html ilustrated by Fig. \ref{LULU1} \normalfont

\itshape Investors are looking for evidence that the company’s product strategy and holiday momentum can drive a sustained recovery. This year and 2025 will be ‘transitional periods,’ said Finance Chief Paula Oyibo. Consumers continue to spend, Oyibo said, but ‘economic concerns are driving a greater focus on value.’ None of our picks is NVIDIA or Amazon or Tesla. Investors may wish to widen the aperture to find the best stocks for outsize earnings growth. Marvell Technology, Lululemon, Block, Amazon and Tesla have been highlighted in this Investment Ideas article. DG reports higher net sales and improved same-store sales in Q3. The company focuses on new store growth and upgrading mature stores for long-term growth. […]
https://finance.yahoo.com/m/b39804b0-0d3e-3317-943b-0197e3e90ebe/lululemon\%E2\%80\%99s-stock-has.html ilustrated by Fig. \ref{LULU2} \normalfont

\section{User Feedback and Pilot Summary}

Upon the conclusion of the four-week pilot study, feedback was solicited from the nine participants through a structured questionnaire and comment while the experiment was ongoing. The objective was to assess the perceived utility, clarity, and overall effectiveness of the LLM-generated briefs in their investment analysis workflow. The feedback received was generally positive and provided crucial insights for future development.

\subsection{Overall Reception and Key Strengths}

The primary value of the system, consistently highlighted by the participants, was its ability to **aggregate and synthesize vast amounts of information** from disparate sources. The briefs were noted for saving a significant amount of time typically spent on data gathering, allowing investors to focus more on interpretation and strategic decision-making. The dynamic, event-driven nature of the daily reports was particularly appreciated, as it allowed users to quickly catch up on the most important developments without having to monitor multiple news outlets.

\subsection{Constructive Feedback and Areas for Improvement}

Alongside the positive reception, participants provided specific, constructive feedback that forms a clear roadmap for enhancing the system. The key themes that emerged were:

\begin{itemize}
    \item \textbf{Enhancing Analytical Depth:} While the breadth of information was praised, some users pointed out areas where the analytical depth could be improved. For instance, specific feedback was given regarding the presentation of quantitative data, such as revenue forecasts for large-cap companies like Amazon (AMZN). This suggests that future iterations should focus on refining the modules responsible for financial projections to deliver more nuanced insights.
    
    \item \textbf{Customization of Delivery and Frequency:} A recurring theme was the desire for greater user control over the information flow. Participants expressed interest in personalizing the service, with questions such as, "I am wondering how often the monitoring would arrive?". This indicates a strong demand for features that would allow users to set their own preferences for the frequency, timing, and potentially even the specific content of the briefs.
\end{itemize}

\subsection{Pilot Summary and Future Directions}

In summary, the pilot study successfully validated the core hypothesis: a RAG-based LLM system can serve as a valuable tool for individual investors by automating data collection and providing timely analytical summaries. The feedback confirms that the system's main strength lies in its time-saving and information-aggregating capabilities.

The constructive criticism received has been instrumental in defining the direction for future work. The next phase of development will focus on two key areas: (1) increasing the sophistication of the quantitative analysis modules to provide deeper, more actionable insights, and (2) building a user preference center to allow for the customization of brief frequency and content. These improvements will be crucial in transitioning the system from a successful proof-of-concept to a practical, indispensable tool for the modern investor.

\section{Conclusions}

This study investigated the practical application of a hybrid system, centered around an LLM in a RAG architecture, to augment the fundamental analysis workflow for individual investors. By generating daily analytical briefs for a diverse portfolio of nine companies, we have demonstrated that such a system can perform a wide range of complex tasks that go far beyond simple data summarization.

Our findings indicate that an AI-augmented system can serve as a powerful co-pilot for investors. The key capabilities demonstrated throughout our four-week pilot study include:
\begin{itemize}
    \item \textbf{Cross-Domain Synthesis:} The system successfully integrated structured quantitative data from financial reports (e.g., Tesla's 10-Q), unstructured real-time news, and high-level macroeconomic indicators (e.g., PMI data) into coherent, multi-layered analytical narratives.
    \item \textbf{Advanced Information Processing:} Beyond simple retrieval, the system demonstrated advanced capabilities such as the ranking of high-volume news flow by investor salience (NVIDIA), the thematic clustering of articles to identify distinct narratives (Lululemon), and the performance of prompt-guided quantitative analysis (Energy Fuels' cash burn rate).
    \item \textbf{Qualitative Reasoning on Unforeseen Events:} The system proved capable of generating relevant, qualitative risk assessments in response to sudden, unpredictable "black swan" events, as shown in the case of the declaration of martial law in South Korea. It was able to tailor its analysis to the specific business models of different companies (Coinbase, NVIDIA, Lululemon), showcasing a degree of situational reasoning.
    \item \textbf{Confirmed Practical Utility:} The feedback from our nine participants confirmed the primary hypothesis of the study. The briefs were consistently valued for their ability to save time and aggregate information, thereby allowing investors to focus on higher-level decision-making.
\end{itemize}

The principal contribution of this work is the holistic, end-to-end demonstration and user-centric evaluation of a hybrid AI system for fundamental analysis. While much research focuses on isolated NLP tasks in finance, our study validates a complete workflow—from heterogeneous data ingestion to the generation of user-vetted analytical products. We also contribute a methodological blueprint that combines the strengths of LLMs for reasoning and synthesis with traditional machine learning (K-Means) and modern data infrastructure (vector databases) for specialized tasks.

Despite the promising results, we acknowledge the limitations of this pilot study. The sample size of nine investors and nine companies over a four-week period means the findings are indicative rather than statistically generalizable. Furthermore, our evaluation focused on the perceived utility of the briefs, not on their direct impact on investment performance; a quantitative backtest of the system's recommendations was outside the scope of this work as our system is being designed as brief generator and advisor, not investment decision system.

In conclusion, as LLMs and related AI technologies continue to mature, their potential to democratize access to sophisticated financial analysis is immense. Our research suggests that hybrid systems, thoughtfully designed and grounded in real-world data, are poised to become indispensable tools for the next generation of investors.
\bibliographystyle{chicago}
\bibliography{sample}

@article{Loughran2011,
    author  = "Tim Loughran and Bill McDonald",
    title   = "When Is a Liability Not a Liability? Textual Analysis, Dictionaries, and 10-{K}s",
    year    = "2011",
    journal = "The Journal of Finance",
    volume  = "66",
    number  = "1",
    pages   = "35--65"
}

@inproceedings{Devlin2019,
    author    = "Jacob Devlin and Ming-Wei Chang and Kenton Lee and Kristina Toutanova",
    title     = "{BERT}: Pre-training of Deep Bidirectional Transformers for Language Understanding",
    year      = "2019",
    booktitle = "Proceedings of the 2019 Conference of the North {A}merican Chapter of the Association for Computational Linguistics: Human Language Technologies, Volume 1 (Long and Short Papers)",
    pages     = "4171--4186"
}

@misc{Araci2019,
    author       = "Dogu Araci",
    title        = "{FinBERT}: Financial Sentiment Analysis with Pre-trained Language Models",
    year         = "2019",
    eprint       = "1908.10063",
    archiveprefix = "arXiv",
    primaryclass  = "cs.CL"
}

@misc{LopezLira2023,
    author  = "Alejandro Lopez-Lira and Yuehua Tang",
    title   = "Can {ChatGPT} Forecast Stock Price Movements? Return Predictability and Large Language Models",
    year    = "2023",
    journal = "SSRN Electronic Journal",
    doi     = "10.2139/ssrn.4412788"
}

@inproceedings{Lewis2020,
    author    = "Patrick Lewis and Ethan Perez and Aleksandra Piktus and Fabio Petroni and Vladimir Karpukhin and Naman Goyal and Yacine Jui and Hema Dwivedi-Yu and Pontus Stenetorp and Sebastian Riedel and Douwe Kiela",
    title     = "Retrieval-Augmented Generation for Knowledge-Intensive {NLP} Tasks",
    year      = "2020",
    booktitle = "Advances in Neural Information Processing Systems",
    volume    = "33",
    pages     = "9459--9474"
}

@misc{Wu2023,
    author       = "Shijie Wu and Ozan Irsoy and Steven Lu and Vadim Dabravolski and Mark Dredze and Sebastian Gehrmann and Prabhanjan Kambadur and David Rosenberg and James Callan",
    title        = "{BloombergGPT}: A Large Language Model for Finance",
    year         = "2023",
    eprint       = "2303.17564",
    archiveprefix = "arXiv",
    primaryclass  = "cs.CL"
}

@misc{Itoh2023,
    author       = "Satoshi Itoh and Katsuhiko Okada",
    title        = "The Power of Large Language Models: A {ChatGPT}-driven Textual Analysis of Fundamental Data",
    year         = "2023",
    howpublished = "Available at SSRN 4425952",
    doi          = "10.2139/ssrn.4425952"
}

@misc{Shuster2021,
      title={{Retrieval Augmentation Reduces Hallucination in Conversation}}, 
      author={Kurt Shuster and Spencer Poff and Moya Chen and Douwe Kiela and Jason Weston},
      year={2021},
      eprint={2104.07567},
      archivePrefix={arXiv},
      primaryClass={cs.CL}
}

@inproceedings{Vaswani2017,
    author    = "Ashish Vaswani and Noam Shazeer and Niki Parmar and Jakob Uszkoreit and Llion Jones and Aidan N. Gomez and Lukasz Kaiser and Illia Polosukhin",
    title     = "Attention Is All You Need",
    year      = "2017",
    booktitle = "Advances in Neural Information Processing Systems 30 (NIPS 2017)",
    pages     = "5998--6008"
}

@misc{OpenAI2024gpt4o,
      title={{GPT-4o}}, 
      author={OpenAI},
      year={2024},
      howpublished={\url{https://openai.com/index/hello-gpt-4o/}},
      note={Accessed: 2025-07-30}
}

@misc{Laniewski2024,
    author       = {Stanislaw Laniewski and Robert \'Slepaczuk},
    title        = {{Enhancing literature review with LLM and NLP methods. Algorithmic trading case}},
    year         = {2024},
    eprint       = {2401.03370},
    archivePrefix = {arXiv},
    primaryClass = {q-fin.CP}
}

@inproceedings{Rzepka2023,
    author    = {Rafał Rzepka and Akihiko Obayashi},
    title     = {{Effectiveness of Security Export Control Ontology for Predicting Answer Type and Regulation Categories}},
    year      = {2023},
    booktitle = {Proceedings of the 21st International Conference on Artificial Intelligence and Law (ICAIL 2023)},
    pages     = {253--257},
    doi       = {10.1145/3594536.3595152}
}

@inproceedings{Rzepka2023perception,
    author    = {Rafa{\l} Rzepka and Kei Okada},
    title     = {{Simulating Perception With LLMs as Underpinnings for More Controllable Knowledge Acquisition}},
    year      = {2023},
    booktitle = {Proceedings of the 2023 IEEE/SICE International Symposium on System Integration (SII)},
    pages     = {1--6},
    doi       = {10.1109/SII55687.2023.10050181}
}

@inproceedings{MacQueen1967,
    author    = {MacQueen, James B.},
    title     = {{Some methods for classification and analysis of multivariate observations}},
    year      = {1967},
    booktitle = {Proceedings of the Fifth Berkeley Symposium on Mathematical Statistics and Probability, Volume 1: Statistics},
    pages     = {281--297},
    publisher = {University of California Press},
    address   = {Berkeley, Calif.}
}

@inproceedings{She2024LanceDB,
    author    = {She, Chang and Brenecki, Leigh and Maddali, Suryanarayana and Le, Tan and Vu, Vy and Jain, Anbuj},
    title     = {{LanceDB - Embracing Composability in the Storage Layer}},
    year      = {2024},
    booktitle = {Proceedings of the VLDB Endowment, Volume 17, No. 13},
    pages     = {3821--3824},
    publisher = {VLDB Endowment},
    doi       = {10.14778/3639368.3639393}
}

@misc{MuellerWordCloud,
    author       = {Andreas Mueller},
    title        = {{word\_cloud: A word cloud generator in Python}},
    howpublished = {\url{https://github.com/amueller/word_cloud}},
    note         = {Accessed: 2025-08-30}
}

\end{document}